\documentclass[10pt,twocolumn,letterpaper]{article}

\usepackage{cvpr}
\usepackage{times}
\usepackage{epsfig}
\usepackage{graphicx}
\usepackage{amsmath}
\usepackage{amssymb}

\usepackage[pagebackref=true,breaklinks=true,letterpaper=true,colorlinks,bookmarks=false]{hyperref}

\cvprfinalcopy 

\def\cvprPaperID{****} 

\ifcvprfinal\fi
\begin{document}

\title{Investigating Bias In Automatic Toxic Comment Detection: An Empirical Study}

\author{Ayush Kumar\\
Georgia Institute of Technology\\
Atlanta, US\\
{\tt\small akumar689@gatech.edu}
\and
Pratik Kumar\\
Georgia Institute of Technology\\
Atlanta, US\\
{\tt\small pkumar319@gatech.edu}
}

\maketitle

\begin{abstract}
With surge in online platforms, there has been an upsurge in the user engagement on these platforms via comments and reactions. A large portion of such textual comments are abusive, rude and offensive to the audience. With machine learning systems in-place to check such comments coming onto platform, biases present in the training data gets passed onto the classifier leading to discrimination against a set of classes, religion and gender. In this work, we evaluate different classifiers and feature to estimate the bias in these classifiers along with their performance on downstream task of toxicity classification. Results show that improvement in performance of automatic toxic comment detection models is positively correlated to mitigating biases in these models. In our work, LSTM with attention mechanism proved to be a better modelling strategy than a CNN model. Further analysis shows that fasttext embeddings is marginally preferable than glove embeddings on training models for toxicity comment detection. Deeper analysis reveals the findings that such automatic models are particularly biased to specific identity groups even though the model has a high AUC score. Finally, in effort to mitigate bias in toxicity detection models, a multi-task setup trained with auxiliary task of toxicity sub-types proved to be useful leading to upto 0.26\% (6\% relative ) gain in AUC scores.
\end{abstract}

\section{Introduction}
With penetration of social media platforms, a growing need to monitor and effectively manage negative behaviors has emerged in the recent past. Moderation is crucial to promoting healthy online discussions. One prime area of focus is to identify textual comments that are abusive, rude and offensive to the audience. Over past few years, several approaches have come up to identify such toxic comments on online platforms to make these interactions free from abuses and hatred. In addition to a binary categorization of toxic vv non-toxic comments, some approaches also identify several categories of toxicity such as profanity, insults, obscenity, threat etc. that to effectively manage the content and user’s behavior on online platforms. Most of these approaches employ representations such as tf-idf or word embeddings to feed the documents to machine learning algorithms. A typical ML algorithm would look like a statistical model such as SVM and random forest or more advanced and state of art models such as convolution neural network, LSTM and transformers. However, some recent works as outlined in Section \ref{related} suggests that toxic classification models tend to classify non-toxic comments into toxic classes that contain some of the commonly attacked identity groups. Google's perspective API has also been reported to generate higher toxicity scores for some of targeted identity groups \footnote{\url{https://medium.com/jigsaw/unintended-bias-and-names-of-frequently-targeted-groups-8e0b81f80a23}}. Higher representation of abusive and toxic comments in the dataset for these identity groups attributed to such bias in the model's predictions. For example, in many forums it’s common to use the word \textit{gay} as an insult, or for someone to attack a commenter for being gay. It is much rarer for the word gay to appear in a positive, affirming statements. These biases are passed on to the trained models via the dataset annotations. Rudimentary machine learning models may learn some unintended bias during the process of training that may be disrespectful against a specific community/identity. In this work, we pose several research questions (RQ) to study the unintended bias in automatic toxic comment detection models:
\begin{itemize}
    \item \textbf{RQ1:} How does the size of the model affect the performance of the toxicity detection system and the associated bias?
    \item \textbf{RQ2:} Does different model architectures lead to varied levels of unintended bias?
    \item \textbf{RQ3:} If we freeze the embedding layer, how does it affect the model performance and the associated bias?
    \item \textbf{RQ4:} Are there specific set of identities against which the model(s) are particularly biased?
    \item \textbf{RQ5:} Does training on auxiliary tasks in addition to toxicity detection helps in mitigating unintended bias?
\end{itemize}

\section{Related Works}
\label{related}
Researchers have already proposed a variety of techniques to reduce biases in the Toxic Comment Detection. Few Researchers tried to show that preprocessing the dataset to mitigate the biases in toxic comment detection is a good approach.  Park et al. (2018) \cite{DBLP:conf/emnlp/ParkSF18} used the combination of debiased word2vec and gender swap data augmentation to reduce the gender bias. Some researchers pay more attention to modifying the models and learning less biased features. Xia et al. (2020) \cite{DBLP:conf/acl-socialnlp/XiaFT20} used adversarial training to reduce the intention of the Toxic Comment Detection system. Researchers \cite{DBLP:conf/aies/SwingerDHLK19} have highlighted how word embeddings exhibit human stereotypes towards genders and ethnic groups. They consider the problem of unsupervised Bias Enumeration (UBE): discovering biases automatically from an unlabeled data representation. Few researchers have proposed to use invariant rationalization,a game-theoretic framework consisting of a rationale generator and predictors, to rule out the spurious correlation of certain syntactic patterns (e.g., identity mentions, dialect) to toxicity labels \cite{DBLP:journals/corr/abs-2106-07240}. They empirically showed that their method yields better lower false positive rate. Such algorithm is useful as it provides way for social scientists to use it as a tool to study human biases and furthermore one can only work towards debiasing only after acknowledging the biases.
Researchers \cite{DBLP:conf/aies/DixonLSTV18} also tried using the  unsupervised approach based on balancing the
training dataset to mitigate the bias in Toxic Comment detection. They got better results and showed that using this method can mitigate the unintended biases in a model without harming the overall model quality. The researchers in \cite{DBLP:conf/icwsm/VaidyaMN20} proposed a multi-task learning model with an attention layer
that jointly learns to predict the toxicity of a comment as well
as the identities present in the comments in order to reduce
this bias.

While some of the works considering discovering and mitigating the biases in text classification including toxicity detection task, our experimental setup and studies are most similar to work done by Vaidya et al. \cite{DBLP:conf/icwsm/VaidyaMN20}. With an overlap in idea of using a multi-task setup to learn a primary and auxiliary task in a joint training paradigm, we additionally probe the experimental setup for on three dimensions: 1. model size, 2. word embeddings, 3. toxicity AUC optimization vs subgroup AUC scores. 

\begin{figure}
\centering
\begin{minipage}{0.45\textwidth}
  \centering
  \includegraphics[width=1.0\linewidth, height=6.0cm]{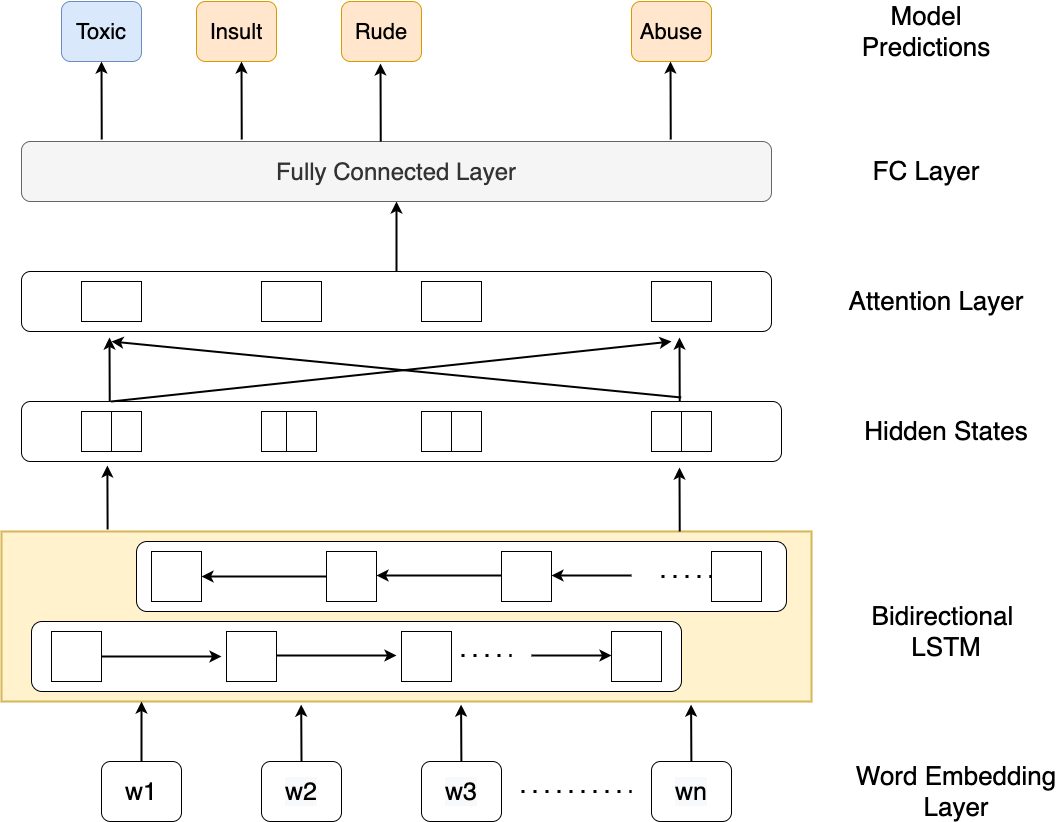}
  \label{setD}
\end{minipage}%
\caption{Overview of architecture diagram for multi-task toxicity prediction using a bidirectional LSTM network with attention.}
\label{exp1}
\end{figure}

\section{Methodology}

In this section, we describe the methodology to arrive at answers to the proposed research questions.

\subsection{Model Architecture}
We experiment with two widely popular architectures for text classification: long-short-term-memory networks (LSTMs) and convolutional neural network (CNNs) as described below:
\begin{itemize}
    \item LSTM: In our work, we use a single layer bidirectional LSTM network that uses a forward LSTM and a backward LSTM to process the input from both directions. The hidden state vector is obtained by concatenating forward ($\vec{h_f}$) and backward ($\overleftarrow{h_b}$) hidden vectors. To obtain forward and backward state, we experiment with three options:
    \begin{itemize}
        \item Final hidden state: In this approach, we simply use the hidden state of last token of the utterance.
        \item Mean and Max pool: In this approach, we concatenate the average and maximum of all hidden states over all tokens in the utterance. This approach is particularly intended towards long utterances where past context information is lost.
        \item Attention Mechanism: Attention is one of the most powerful strategy in deep learning models to identify the most significant features (tokens in an utterance) that is helpful for the downstream task. In this work, we use compute attention score ($a_i$) for each token in the utterance \cite{DBLP:conf/emnlp/ZhouWX16}. This attention score is used as for weighted combination of hidden states ($h_u$):
        \begin{equation}
            a_i = \frac{exp(tanh(W_ah_i))}{\sum_{j=1}^{n}exp(tanh(W_ah_j))}
        \end{equation}
        
        \begin{equation}
            h_u = \sum_{i}^{n}a_ih_i
        \end{equation}
    \end{itemize}
    
    \item CNN: We use three layer 1d-CNN model \cite{DBLP:conf/emnlp/Kim14} i.e, the model contains three convolution + pooling layer before the feeding the representations to feed forward layer for final predictions. Max-pool operation is used at pooling layer to aggregate features from different filters. 
\end{itemize}

\subsection{Multi-task Setup}
In addition to training a classifier network for toxcitiy comment classification, we also predict toxicity sub-types present in the dataset:  \textit{severe toxicity, obscene, threat, identity attack,} and \textit{insult}. To train such model, we share the utterance input and representation layer keeping different classification layer for the two tasks at hand. The rationale behind such setup is that to improve the accuracy of detecting toxic comments along with predicting toxicity sub-types. Joint learning setup can act as inherent regularizer for the other tasks due to share representation layer making a multi-task setup a competitive choice for training the model for toxicity classification.

\subsection{Utterance Embedding}
Each utterance $u$ is represented as a matrix of $t \times d$ vectors, where $t$ is the number of tokens in the utterance and $d$ is the embedding dimension. We use two pre-computed embedding lookups from glove vectors \cite{pennington2014glove} and fasttext embeddings \cite{grave2018learning} to generate document-embeddings fed to the classification network at the input layer. In the training setup, we evaluate the impact of frozen embeddings against making the embedding layer trainable.

\subsection{Model Prediction and Loss Function}
The primary downstream task at hand is toxic comment classification which is a binary classification task. The prediction layer $t$ for binary classification task can be represented as:

\begin{equation}
    y_t = \sigma(w_t^T h + b_t)
\end{equation}

Additionally, in multi-task setup, we have toxicity sub-type classification as auxiliary task where model needs to predict the toxicity sub-types across six labels with one or more labels being correct. Thus, toxicity sub-type classification is a multi-label classification task which can be represented as:

\begin{equation}
    y_s^k = \sigma(w_s^T h + b_s), k=1,2,...,K
\end{equation}

Since we have a binary and multi-label classification at hand, we use a cross-entropy (CE) loss:
\begin{equation}
    L = \sum_i^{C} {\hat{y}\log(y_t) + (1 - \hat{y})\log(1 - y_t)}
\end{equation}

\subsection{Evaluation Metrics}
For primary toxicity comment detection task, we use ROC-AUC as the evaluation metric, referred as Toxicity AUC from hereon. In addition to this, to evaluate the models against the proposed research questions, we use three sub-metrics as per the task evaluation framework \footnote{\url{https://www.kaggle.com/c/jigsaw-unintended-bias-in-toxicity-classification/overview/evaluation}}:

\begin{itemize}
    \item \textbf{Subgroup AUC:} To calculate this metric, the test set is restricted to only those datapoints that contains a particular identity subgroup. A lower subgroup AUC value would mean that the model does a poor job at distinguishing the toxic vs non-toxic comments wherever specific identity is mentioned.
    
    \item \textbf{BPSN AUC:} BPSN or Background Positive, Subgroup Negative represents those datapoints where the comments are toxic while subgroup is absent or comments are non-toxic while identity subgroup is mentioned in the comment. A low BPSN AUC value signifies the case where model confuses the non-toxic comments that mentions identity with toxic comments that don't. In other words, model predicts higher toxicity score than it should for the comments that mentions an identity subgroup.
    
    \item \textbf{BNSP AUC:} For BNSP or Background Negative, Subgroup Positive AUC, the test set is restricted to the datapoints where toxic comments mentions the identity while non-toxic comments do not. A low BNSP AUC would refer to the scenario where model predicts lower scores for the toxic comments containing identity subgroup than it should. 
\end{itemize}

The final evaluation metric, called as \textit{generalized AUC} is measured as:

\begin{equation}
    M_p(m_s) = (\frac{1}{N} \sum_{s=1}^{N}m_s^p)^\frac{1}{p}
\end{equation}
where, $M_p$ is called as pth power mean function, $m_s$ is subgroup bias metric for identity group $s$ and $N$ is the number of identity groups. A value of p=5 is choosen as per the task evaluation methodology which enourages to focus on subgroup bias in addition to toxicity AUC scores.

\section{Experimental Setup}

\subsection{Dataset and Preprocessing}

In this work, we use dataset provided in \textit{Jigsaw Unintended Bias in Toxicity Classification} Kaggle competition. The challenge is designed to build a model that recognizes toxicity and minimizes unintended bias with respect to mentions of identities. The dataset contains more than 1.8M datapoints each annotated by upto 10 annotators. The toxicity score for any comment is determined by the fraction of annotators who have marked a comment as toxic. Each comments was further marked with toxicity sub-types under the categories: \textit{severe toxicity, obscene, threat, identity attack, and insult}.

In addition to toxicity labels, annotators are also asked to label identity subgroups mentioned in the dataset. A sample annotation from the corpus looks like this:
\noindent\vspace{\baselineskip}

Comment: \textit{Continue to stand strong LGBT community. Yes, indeed, you'll overcome and you have.}\\
\textit{Toxicity Labels:} All 0.0\\
\textit{Identity Mention Labels:} homosexual\_gay\_or\_lesbian: 0.8, bisexual: 0.6, transgender: 0.3 (all others 0.0)
\noindent\vspace{\baselineskip}

As per the competition guidelines, we binarize all labels by assigning a label of 1 wherever $score >= 0.5$. The dataset distribution across identities is mentioned in Table \ref{data}. In the preprocessing step, we remove all special (non-alphanumeric) characters and emoticons. We also truncate any comment to a maximum of 200 sequence length.

\begin{table}[]
\centering
\renewcommand*{\arraystretch}{1.1}
\begin{tabular}{lcc}
\multicolumn{1}{c}{\textbf{Identities}} & \textbf{Count} & \textbf{Toxic \%} \\ \hline
black                                   & 17,161         & 19.69\%           \\
white                                   & 28,831         & 17.76\%           \\
male                                    & 64,544         & 9.60\%            \\
female                                  & 55,048         & 9.14\%            \\
homosexual (gay	or	lesbian)             & 11,060         & 19.01\%           \\
christian                               & 40,697         & 5.59\%            \\
muslim                                  & 21,323         & 14.94\%           \\
jewish                                  & 7,669          & 10.25\%           \\
psychiatric or	mental	illness           & 6,218          & 14.09\%           \\
All Identities                          & 191,671        & 14.44\%          
\end{tabular}
\caption{Distribution of toxic comments across identity sub-groups. }
\label{data}
\end{table}

\subsection{Model hyperparameters and training}
We conduct experiments with varying hyperparameters. For model type we used $CNN$ and $LSTM$ which are state of the art for many text classification tasks (prior to transformers era). To study the impact of model size on the performance, we identify the architecture on 3 sizes $S$,$M$,$L$ depending on the number of hidden layers/filters of the model. For $S$, we use $64$ hidden layers/filters in the LSTM/CNN model. For $M$ and $L$, we use $256$ and $512$ hidden layers/filters in the LSTM/CNN model respectively.  We use two embedding models $glove$ and $fasttext$ to understand the dependency of the embedding vectors on the bias. We also experimented with the freezing the embedding layer unlike the most suggested option of fine-tuning the embedding. We train them in half the experiments and we froze them in half the experiments to see the biases in the pretrained embedding model. We perform the experiments for 5 epochs each and use early stopping to choose the model with best validation loss. We stop training the model if validation loss does not improve over most recent three epochs. For optimization, we use cross entropy loss as specified in Eqn. 5 trained with Adam optimizer with a learning rate choosen over {1e-3, 1e-4 and 1e-5} for respective models. We code the model in Pytorch framework and the model is trained on a K-80 GPU.

For model training, we do not use any code repository directly. However, our \hyperlink{https://www.kaggle.com/dborkan/benchmark-kernel}{evaluation scripts} are inspired from the available codes on Kaggle platform.

\section{Results and Analysis}

\begin{table*}[]
\centering
\renewcommand*{\arraystretch}{1.1}
\begin{tabular}{cccccc}
Model & Model Size & Word Embedding & Trainable & AUC Score & Bias Metric \\ \hline
LSTM  & S          & fasttext       & False     & 0.9592    & 0.9278      \\
      &            & fasttext       & True      & 0.9615    & 0.9317      \\
      &            & glove          & False     & 0.9577    & 0.9259      \\
      &            & glove          & True      & 0.9622    & 0.9323      \\
      &            &                &           &           &             \\
      & M          & fasttext       & False     & 0.9605    & 0.9303      \\
      &            & fasttext       & True      & 0.9546    & 0.9253      \\
      &            & glove          & False     & 0.9598    & 0.9296      \\
      &            & glove          & True      & 0.9627    & 0.9325      \\
      &            &                &           &           &             \\
      & L          & fasttext       & False     & 0.9612    & 0.9305      \\
      &            & fasttext       & True      & 0.9627    & \textbf{0.9347}      \\
      &            & glove          & False     & 0.9611    & 0.9303      \\
      &            & glove          & True      & \textbf{0.9634}    & 0.9346      \\
      &            &                &           &           &             \\
CNN   & S          & fasttext       & False     & 0.9549    & 0.9228      \\
      &            & fasttext       & True      & 0.9601    & 0.9289      \\
      &            & glove          & False     & 0.9518    & 0.9200        \\
      &            & glove          & True      & 0.9605    & 0.9283      \\
      &            &                &           &           &             \\
      & M          & fasttext       & False     & 0.9555    & 0.9237      \\
      &            & fasttext       & True      & 0.9597    & 0.9275      \\
      &            & glove          & False     & 0.9535    & 0.9190       \\
      &            & glove          & True      & 0.9600      & 0.9271      \\
      &            &                &           &           &             \\
      & L          & fasttext       & False     & 0.9568    & 0.9267      \\
      &            & fasttext       & True      & \textbf{0.9614}    & \textbf{0.9296}      \\
      &            & glove          & False     & 0.9545    & 0.9228      \\
      &            & glove          & True      & 0.9612    & 0.9293     \\
      &            &           &       &     &      \\
SOTA \cite{DBLP:conf/icwsm/VaidyaMN20}    &     -       & glove+fasttext          & True      & 0.9709    & 0.9407     \\
    &            &           &       &     &      \\
\end{tabular}
\caption{AUC scores on toxicity comment classification task for LSTM and CNN models trained with fasttext and glove embeddings.}
\label{results}
\end{table*}

\subsection{RQ1-3: Model Size, Architecture, Embedding}
Table \ref{results} present the results pertaining to research questions on how does model size, architecture and frozen embedding layer affects the model performance and the unintended bias towards identity groups. Based on results, here are the key observations:
\begin{itemize}
    \item Model Size: Model with largest parameters i.e, model size L leads to best performance for both model architectures (LSTM: 0.9634 and CNN: 0.9614). Additionally, model L has best bias metric results too (LSTM: 0.9346 and CNN: 0.9296).
    
    \item Model Architecture: LSTM model with attention leads to better results as compared to CNN for both toxic classification and the associated bias. However, the difference is marginal as evident from detailed results in Table \ref{results}.
    
    \item Frozen Embedding: The embedding layer, if frozen, leads to sub-optimal results for each respective comparisons on word embedding (fasttext, glove) or model sizes. On average statistics, freezing embedding layer downgrades the model performance by 0.36\% on AUC metric and 0.48\% on bias metric. On a deeper look, the degradation is higher (0.6\%) for CNN models as compared to LSTM models.
\end{itemize}

\begin{table*}[]
\centering
\renewcommand*{\arraystretch}{1.1}
\begin{tabular}{l|cccc|c}
\multicolumn{1}{c|}{\textbf{Identity Subgroup}} & \textbf{\begin{tabular}[c]{@{}c@{}}LSTM - L, \\ fasttext\end{tabular}} & \textbf{\begin{tabular}[c]{@{}c@{}}LSTM - L, \\ glove\end{tabular}} & \textbf{\begin{tabular}[c]{@{}c@{}}CNN - L, \\ fasttext\end{tabular}} & \textbf{\begin{tabular}[c]{@{}c@{}}CNN - L, \\ glove\end{tabular}} & \textbf{Avg. AUC} \\ \hline
black                                           & \textbf{0.8731}                                                                 & 0.8663                                                              & 0.8454                                                                & 0.8465                                                             & 0.8578            \\
white                                           & \textbf{0.8916}                                                                 & 0.8879                                                              & 0.8806                                                                & 0.8803                                                             & 0.8851            \\
male                                            & \textbf{0.9463}                                                                 & 0.9435                                                              & 0.9377                                                                & 0.9367                                                             & 0.9410            \\
female                                          & \textbf{0.9504}                                                                 & 0.9481                                                              & 0.9438                                                                & 0.9431                                                             & 0.9463            \\
homosexual\_gay\_or\_lesbian                    & 0.8626                                                                 & 0.8620                                                              & \textbf{0.8628}                                                                & 0.8613                                                             & 0.8622            \\
christian                                       & \textbf{0.9618}                                                                 & 0.9609                                                              & 0.9608                                                                & 0.9617                                                             & 0.9613            \\
jewish                                          & 0.9365                                                                 & \textbf{0.9376}                                                              & 0.9235                                                                & 0.9296                                                             & 0.9318            \\
muslim                                          & \textbf{0.9111}                                                                 & 0.9094                                                              & 0.8976                                                                & 0.8995                                                             & 0.9044            \\
psychiatric\_or\_mental\_illness                & 0.9179                                                                 & 0.9175                                                              & \textbf{0.9239}                                                                & 0.9197                                                             & 0.9198            \\ \hline
\multicolumn{1}{c|}{\textbf{Avg. AUC}}          & 0.9168                                                                 & 0.9148                                                              & 0.9085                                                                & 0.9087                                                             &                  
\end{tabular}
\caption{Subgroup AUC}
\label{subgroup}
\end{table*}

\begin{figure}
\centering
\begin{minipage}{0.48\textwidth}
  \centering
  \includegraphics[width=1.0\linewidth, height=5.0cm]{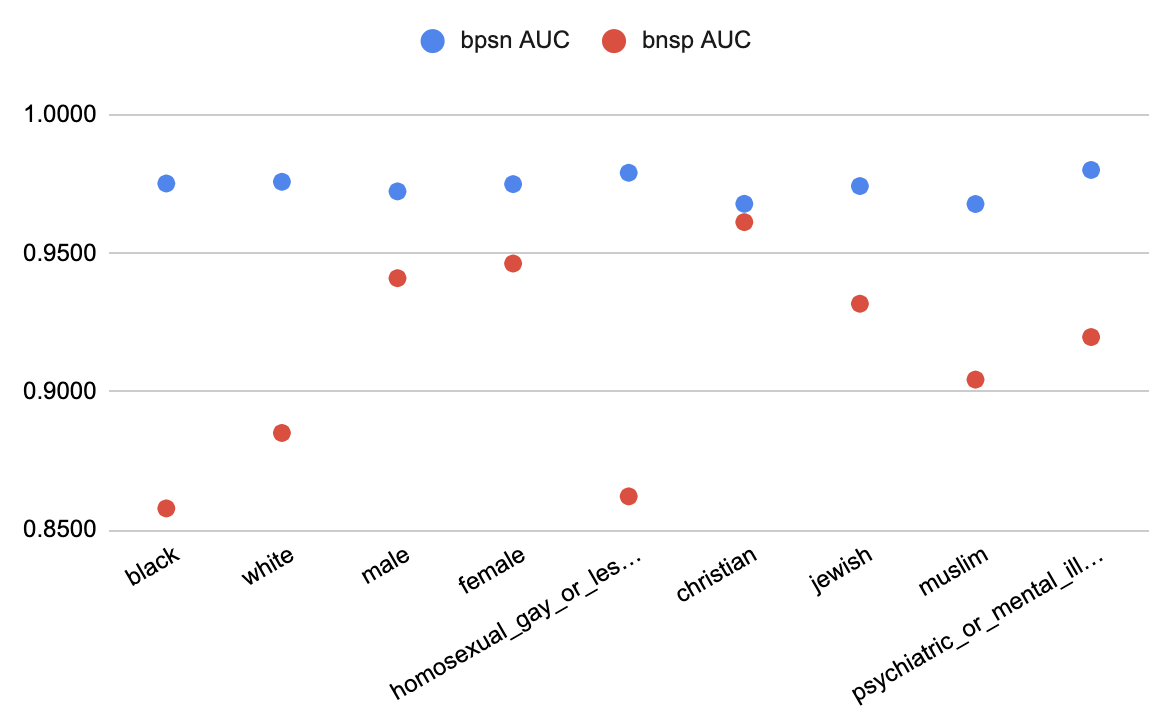}
  \label{setD}
\end{minipage}%
\caption{Tradeoff between BPSN AUC and BNSP AUC scores for different identity groups.}
\label{tradeoff}
\end{figure}

\begin{table}[]
\centering
\begin{tabular}{lccc}
\renewcommand*{\arraystretch}{1.1}
\textbf{Model}                              & \textbf{\begin{tabular}[c]{@{}c@{}}Toxicity \\ AUC\end{tabular}} & \textbf{\begin{tabular}[c]{@{}c@{}}Avg. Subgroup \\ AUC\end{tabular}} & \textbf{\begin{tabular}[c]{@{}c@{}}Generalized \\ AUC\end{tabular}} \\ \hline
\textbf{LSTM}                               & 0.9627                                                           & 0.8827                                                                & 0.9347                                                              \\
\multicolumn{1}{r}{\textit{\textbf{+ MTL}}} & \textit{\textbf{0.9653}}                                         & \textit{\textbf{0.8841}}                                              & \textit{\textbf{0.9359}}                                            \\
\textbf{}                                   &                                                                  &                                                                       &                                                                     \\
\textbf{CNN}                                & 0.9614                                                           & 0.8662                                                                & 0.9296                                                              \\
\multicolumn{1}{r}{\textit{\textbf{+ MTL}}} & \textit{\textbf{0.9628}}                                         & \textit{\textbf{0.8770}}                                              & \textit{\textbf{0.9310}}    \\
\textbf{}                                   &                                                                  &                                                                       &                                                                     \\
\end{tabular}
\caption{Impact of MTL on model performance}
\label{mtl}
\end{table}

While it is not strictly consistent, the results also show that an increase in model performance for toxicity comment classification also leads to an improvement in bias metric with a few exceptions. Out of top five models selected purely on the basis of AUC score, except for top two ranks, the models stack in the same order on bias metric as well. One rationale for this correlation could be the fact that generalized AUC computation factors on AUC score and thus an increase in AUC leads to improvement in generalized AUC. Decoupling this metrics may be a future task that could be looked into.

\subsection{RQ4: Bias towards identity subgroups}
Since dataset comes with annotation for mention of identity subgroups, we measure subgroup specific biases as reported in Table \ref{subgroup} for nine identity groups that contain more than 500 examples in the test set (as specified in the competition rules). The subgroup identities at hand are: \textit{black, white, make, female, homosexual, christian, jewish, muslim, psychiatric}. When the subgroup AUC are calculated for these identities, following observations could be gathered:
\begin{itemize}
    \item \textit{black} and \textit{homosexual} are the identity groups having the highest bias from the models. Along with that, even \textit{white} ranks among the identity groups facing the most bias from automatic toxicity detection.
    
    \item Toxicity detection models are able to identify toxicity for \textit{male, female, christian} identities with considerably less bias.
    
    \item Embedding source be it glove or fasttext does not lead to distinctive results overall as the scores obtained by the respective models are very similar to each other (Avg. AUC per model in Table \ref{subgroup}). However, take a granular look at the results, we observe that \textit{fasttext} embedding leads to less biased models in 8 out of 9 identity groups.
    
    \item Almost all identity groups have consistent BPSN AUC scores but BNSP AUC scores vary significantly for different groups (Figure \ref{tradeoff}). \textit{black, white, homosexual} have particular a widened gap between the two scores suggesting that model predicts lower toxicity scores for these groups.
\end{itemize}

\subsection{RQ5: Multi-task training with auxiliary task}
\begin{figure}
\centering
\begin{minipage}{0.48\textwidth}
  \centering
  \includegraphics[width=1.0\linewidth, height=5.0cm]{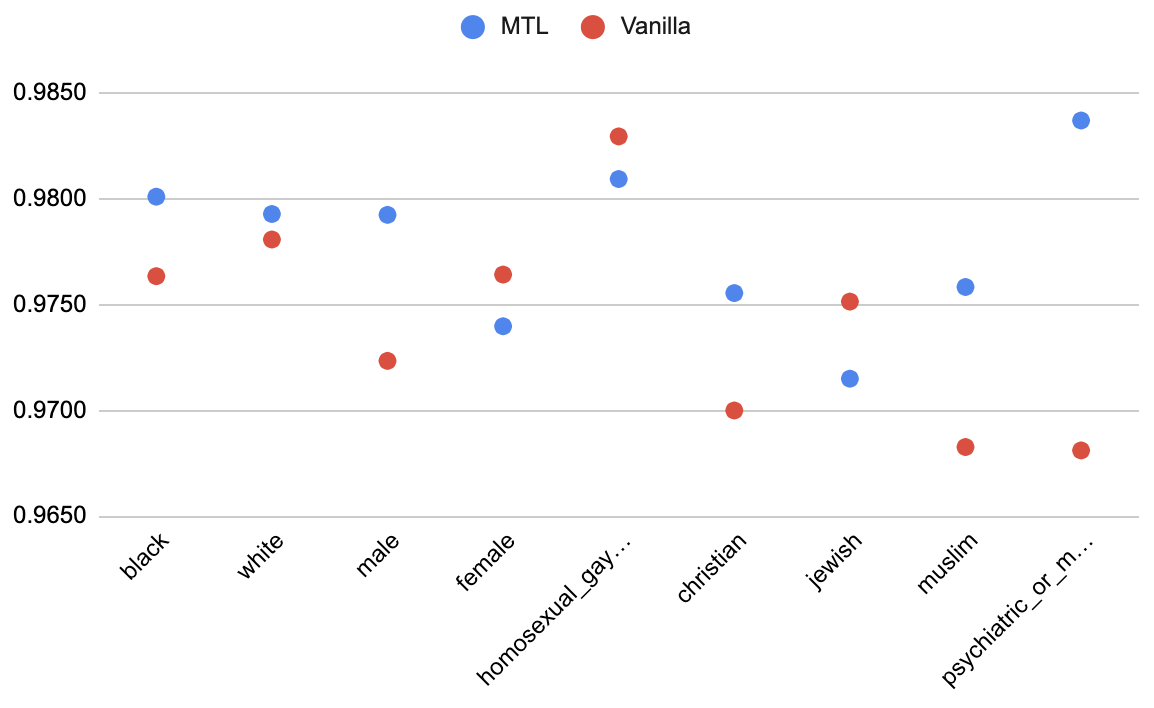}
  \label{setD}
\end{minipage}%
\caption{Impact of MTL training on BPSN AUC scores across identity labels.}
\label{mtl_bpsn}
\end{figure}

In an effort to mitigate unintended bias in the automatic toxicity detection systems, we attempt to leverage the toxicity sub-types annotations to train a multi-task system that jointly learns toxicity binary as well as sub-type labels with a shared representation layer. The results are reported in Table \ref{mtl}. The observations based on the reported results are:

\begin{itemize}
    \item Multi-task learning setup distinctively leads to better model performance across all reported metrics inclusive of toxicity AUC, subgroup AUC and generalized AUC.
    
    \item The reported gain is 0.26\% on toxicity AUC, 0.14\% on avg. subgroup AUC and 0.12\% on generalized AUC metrics for LSTM model.
    
    \item Both LSTM and CNN model benefits from the multi-task setup suggesting that such methods could be a potential improvement over vanilla models that could additionally help mitigate unintended model bias.
    
    \item Figure \ref{mtl_bpsn} shows the impact of multi-task training on BPSN AUC score which reflects the false positivity ratio. The graph demonstrates that MTL training helps the model reduce false positivity ratio on six out of nine identity groups. For some of the identity groups such as make, christian, psychiatric, this gap is wide enough to consider MTL contributing to significant improvement in reduction of FP rate for these groups.
\end{itemize}

\section{Conclusions and Future Works}
Based on the findings in this work, we note that there is a high correlation in a better toxicity comment detection performance and lower bias in the model. We evaluate the models across 3 dimensions: model size, architecture and embeddings. LSTM with attention mechanism turned out to be the best model, both wrt performance on toxicity classification as well as subgroup AUC and generalized AUC. However, there are a few identity groups against which models are particularly biased which needs to be taken a deeper look at. In future, we would want to explore the effectiveness of recent transformers based state-of-art models and understand how large language models fare in such tradeoffs. In addition to this, we would also like to train and evaluate the model on a multi-task setup trained on identity prediction as an auxiliary task.

\section{Ethical Implications}
We note that such automatic toxicity detection model could potentially lead to biased results towards specific identity groups. In light of this, we do not release any of our trained models in public. Additionally, it is in no intention of ours to promote building better models by completely ignoring such societal impacts. Our work in the manuscript is an attempt to understand the unintended bias in toxicity detection systems and identify ways to mitigate such bias. We believe that in areas of such sensitive application, unintended bias in the automatic

{\small
\bibliographystyle{ieee_fullname}
\bibliography{egbib}
}

\end{document}